\documentclass{article}
\usepackage{DS-FONT}

\usepackage{times}
\usepackage{soul}
\usepackage{url}
\usepackage[hidelinks]{hyperref}
\usepackage[utf8]{inputenc}
\usepackage[small]{caption}
\usepackage{graphicx}
\usepackage{amsmath}
\usepackage{amsthm}
\usepackage{booktabs}
\usepackage{algorithm}
\usepackage{algorithmic}
\usepackage{booktabs}
\usepackage{amsmath}
\usepackage{algorithm}
\usepackage{algorithmic}
\usepackage{tabularx}
\usepackage{amsfonts}
\usepackage{float}
\usepackage{subcaption}
\usepackage{longtable}
\usepackage{appendix}
\usepackage{multirow}

\title{Few-shot Font Generation by Learning Style Difference and Similarity}

\author{
	Xiao He$^{1}$
	\and
	Mingrui Zhu$^{1}$\and
	Nannan Wang$^{1}$ \footnote{Co-corresponding authors} \and
    Xinbo Gao$^{1,2}$  \And
    Heng Yang$^{3}$
	\affiliations
	$^1$State Key Laboratory of Integrated Services Networks, Xidian University\\
	$^2$Chongqing Key Laboratory of Image Cognition, Chongqing University of Posts and Telecommunications\\
	$^3$Shenzhen AiMall Tech
}

\begin{document}

\maketitle

\begin{abstract}
		
Few-shot font generation (FFG) aims to preserve the underlying global structure of the original character while generating target fonts by referring to a few samples. It has been applied to font library creation, a personalized signature, and other scenarios. Existing FFG methods explicitly disentangle content and style of reference glyphs universally or component-wisely. However, they ignore the difference between glyphs in different styles and the similarity of glyphs in the same style, which results in artifacts such as local distortions and style inconsistency. To address this issue, we propose a novel font generation approach by learning the Difference between different styles and the Similarity of the same style (DS-Font). We introduce contrastive learning to consider the positive and negative relationship between styles. Specifically, we propose a multi-layer style projector for style encoding and realize a distinctive style representation via our proposed Cluster-level Contrastive Style (CCS) loss. In addition, we design a multi-task patch discriminator, which comprehensively considers different areas of the image and ensures that each style can be distinguished independently. We conduct qualitative and quantitative evaluations comprehensively to demonstrate that our approach achieves significantly better results than state-of-the-art methods.
\end{abstract}

\begin{figure}[t]
\centering
\includegraphics[width=1.0\linewidth]{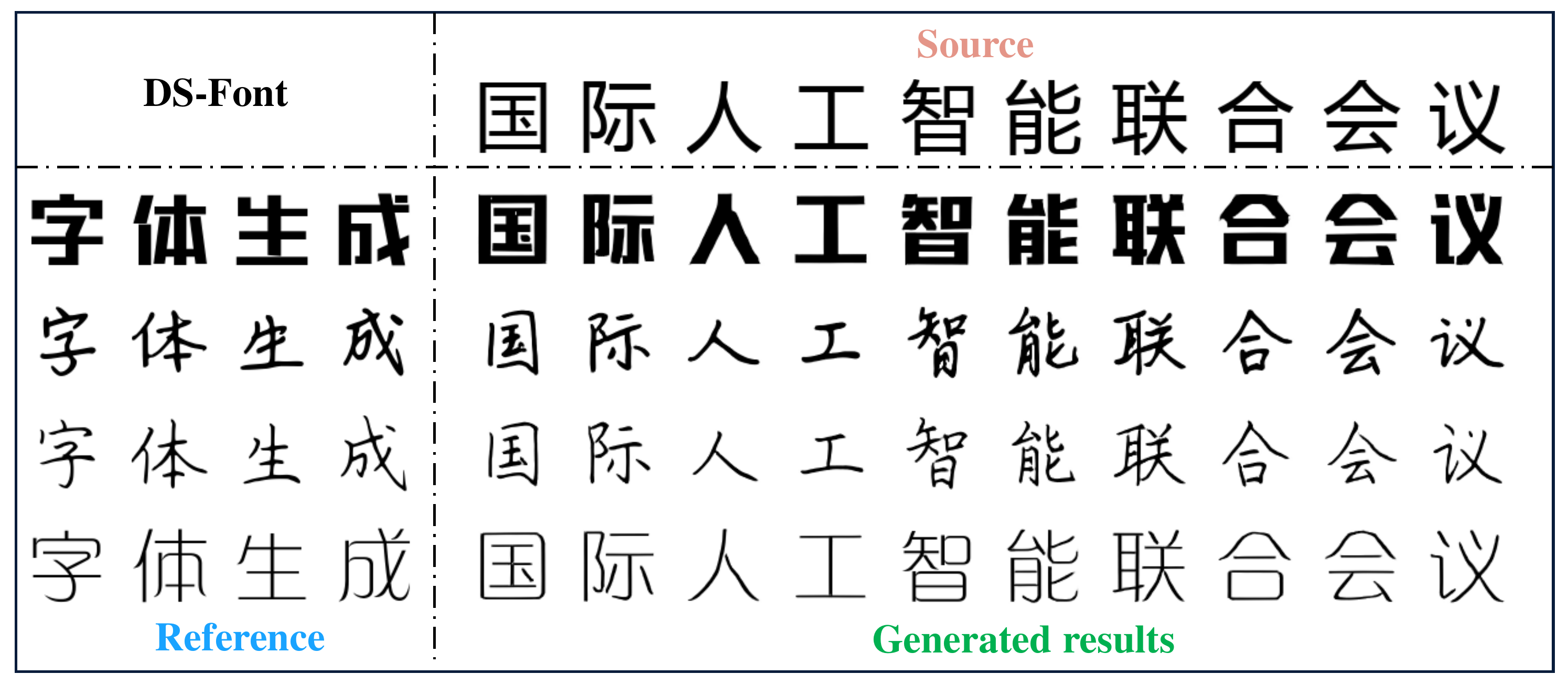}
\caption{\textbf{Font generation results of DS-Font.} The Chinese character on the left is the style image, and the Chinese character on the top right is the content image. The rest is the image generated by our proposed DS-Font. The proposed DS-Font preserves the underlying global structure of the original character while generating target fonts by referring to a few samples.
}
\label{fig:intro}
\end{figure}

\section{Introduction}
In the modern era, text information is a basic form of multimedia information closely related to our life. The font is used as the visual interpretation of words and plays an increasingly important role in the Internet and people's daily life. Font generation is critical in many applications, $e.g.,$ the design of personal signatures, the generation of font libraries, optical character recognition, and the reproduction of historical characters. It is necessary to find an effective and economical way to generate fonts.

Traditional font generation methods rely heavily on expert design and need to skillfully use tools to draw each glyph, which is especially expensive and labor-intensive. The cost is even higher for languages with a large number of characters (Japanese, Chinese characters, Korean, Thai, etc.).

\begin{figure*}[t]
\centering
\includegraphics[width=1.0\linewidth]{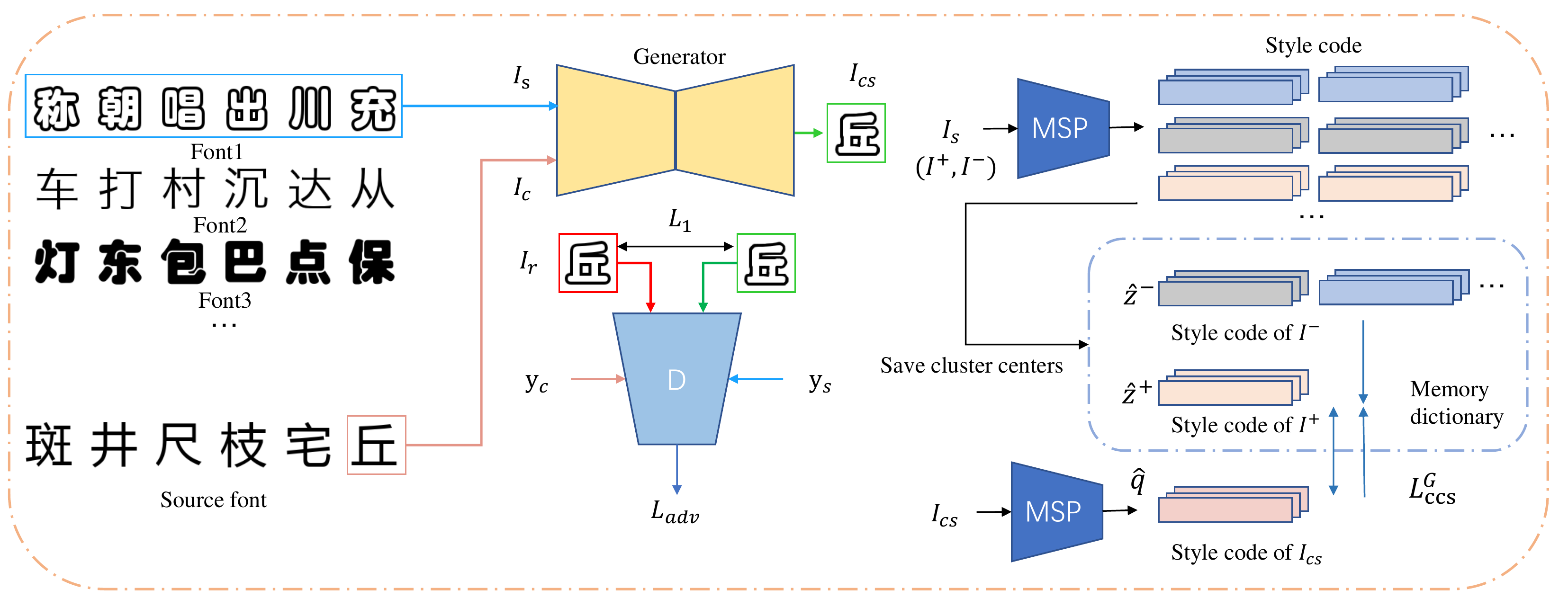}
\caption{\textbf{Overview of DS-Font.} DS-Font consists of a multi-layer style projection (MSP) module, a multi-task patch discriminator, and a generator based on an attention mechanism. We first feed all the images into the MSP module to generate the corresponding style code, calculate the class center of each style according to the style label, and store the code $\hat{z}_j$ in the memory dictionary. (Note that the style images $I_s$ contain images $I^+$ with the same style as the generated image and images $I^-$ with different styles from the generated image.) Then we generate images $I_{cs}$ from a content image $I_c$ and a few style images $I_s$ and feed $I_{cs}$ into the MSP module to get style code as query $\hat{q}$. The style code $\hat{z}^+$ of the image $I^+$ mapped by the MSP module is used as a positive sample. Similarly, the style code $\hat{z}^-$ of the image $I^-$ is used as a negative sample. We calculate cluster-level contrastive style loss $\mathcal{L}_{contra}^{G}$ according to these style codes. Adversarial loss $\mathcal{L}_{adv}$ is based on discriminator. L1 loss $\mathcal{L}_{1}$ calculation generates the pixel-level difference between the generated image and ground truth.  }
\label{fig:meth_arch}
\end{figure*}

With the development of deep learning and the enhancement of hardware computing power, automatic font generation technology has attracted the attention of many computer vision researchers. Automatic font generation technology needs to meet two requirements: 1) the generated font should maintain the basic global structure of the original font. 2) the generated font should present the same style pattern as the target font. Some methods  \cite{chang2018chinese,gao2020gan,hassan2021unpaired,sun2018pyramid,tian2017zi2zi}  based on Generative Adversarial Networks (GAN) \cite{goodfellow2020generative,mirza2014conditional} try to meet these requirements and generate reasonable glyphs. Early attempts, such as zi2zi \cite{tian2017zi2zi}, used a network similar to Pix2Pix  \cite{isola2017image} with font category embedding conditions to learn multiple font styles of a single model. However, these methods need to be pre-trained on large datasets and then fine-tuned for specific tasks. This procedure is time-consuming.

In recent years, some efforts \cite{zhang2018separating,wen2021zigan,gao2019artistic,li2021few,xie2021dg,cha2020few,park2021few,park2021multiple,zhang2022mf}  have attempted to generate fonts using only a few samples. They regard the characters in the standard font library as content glyphs and transfer them to the target font by extracting style representation from the reference glyphs. According to the different forms of style representation, we divide the few-shot font generation methods into two categories: universal style representation and component-wise style representation. The former \cite{liu2019few,gao2019artistic,xie2021dg} models the style of each font as a universal representation, while the latter \cite{cha2020few,park2021few,park2021multiple} decomposes glyphs into many sub-components to extract component-wise style representation.

However, in the font library, the style of each font is hard to define and represent. The detailed styles of components, strokes, and even edges in the same style should be consistent. It is unreasonable for existing methods to divide component-wise or universal styles explicitly. At the same time, because the content and style are highly entangled, it is difficult to ensure the consistency of component-wise styles between the reference glyphs and the generated glyphs by using explicit disentanglement. Thus, some efforts \cite{li2021few,zhang2022mf,tang2022few} attempt to use an attention mechanism to capture multi-level style patterns. However, they ignore the difference between glyphs in different styles and the similarity of glyphs in the same style, which results in artifacts such as local distortions and style inconsistency. Based on the complexity of font styles, our key insight is: if only a few glyphs are given, it is difficult for a person without artistic knowledge to define their writing style, but it is relatively easy to identify whether their writing styles are the same. Therefore, we propose to learn style representation from image features by analyzing the relationship among multiple styles rather than extracting universal or component-wise style representation.

In this paper, we propose a novel FFG method dubbed as DS-Font. Its core idea is to learn style representation from image features by analyzing the difference between different styles and the similarity of the same style. We employ contrastive learning to model the relationship between different styles. It pulls the same style closer and the different styles away. Specifically, we propose a multi-layer style projector for style encoding and realize a distinctive style representation under the supervision of our designed Cluster-level Contrastive Style (CCS) loss. In addition, we design a multi-task patch discriminator, which comprehensively considers different areas of the image and ensures that each style can be distinguished independently. Combing these components, we have achieved high-quality font generation (as shown in Figure \ref{fig:intro}).

Our contributions can be summarized as follows:
\begin{itemize}
\item  We rethink the style representation of glyphs and propose a novel few-shot font generation method instead of using universal or component-level style representation.

\item  An MSP module for style coding is proposed, and a cluster-level contrastive style loss is introduced for supervision, which solves the problem that the existing font generation models can not make full use of a large amount of style information. In addition, in order to better distinguish different styles, a multi-task patch discriminator is proposed.

\item  By considering the relationship among multiple fonts, our approach achieves significantly better results than state-of-the-art methods.

\end{itemize}

\section{Related Works}
\subsection{Image-to-Image Translation}
Image-to-image (I2I) translation methods \cite{choi2020stargan,liu2019few,zhu2017deep,zhang2019neural,zhang2019deep} aim to learn the mapping function from the source domain to the target domain. In recent years, image translation methods have made significant improvements in multi-mapping \cite{choi2020stargan,yu2019multi}, few-shot learning \cite{liu2019few}, and high-resolution processing\cite{anokhin2020high}. For unpaired data, CycleGAN \cite{zhu2017unpaired} proposes a cycle consistency constraint, which can preserve the content structure of the source image during cross-domain translation. FUNIT \cite{liu2019few} encodes content image and style image, respectively, and combines them with adaptive instance normalization (AdaIN) \cite{huang2017arbitrary} to accomplish the I2I translation task. Intuitively, font generation is a typical I2I translation task, which maps the source font into the target font while preserving the original character structure. Thus, many font generation methods are based on I2I translation methods.
\subsection{Many-shot Font Generation}
Early font generation methods \cite{chang2018chinese,gao2020gan,hassan2021unpaired,sun2018pyramid,tian2017zi2zi,jiang2017dcfont,jiang2019scfont} aim to train the mapping function between source fonts and target fonts. A number of font generation methods first train a translation model and fine-tune the translation model with many reference glyphs, $e.g.,$ 775 \cite{jiang2019scfont}. Despite their remarkable performance, their scenario is not practical because collecting hundreds of glyphs with a coherent style is too expensive. In this paper, we aim to generate an unseen font library without expensive fine-tuning or collecting a large number of reference glyphs for a new style.
\subsection{Few-shot Font Generation}
Since the style of glyphs is highly complex and fine-grained, utilizing global feature statistics to transfer the font is difficult. Instead, most few-shot font generation methods try to disentangle content and style features from the given glyphs. Based on different kinds of feature representation, we categorize existing FFG methods into two main categories: universal feature representation \cite{zhang2018separating,wen2021zigan,gao2019artistic,xie2021dg}, and component-based feature representation \cite{cha2020few,park2021few,park2021multiple}. The universal feature representation methods, such as EMD \cite{zhang2018separating} and AGIS-Net \cite{gao2019artistic}, synthesize a new glyph by combining the style feature encoded from the reference set and the content feature encoded from the source glyph. DG-Font \cite{xie2021dg} proposes feature deformation skip connection based on the unsupervised method TUNIT \cite{baek2021rethinking}. Studies related to component-based feature representation focus on decomposing glyphs according to the structure and obtaining multiple localized style representations. LF-Font \cite{park2021few} obtains component-wise style features by designing a component-based style encoder. MX-Font \cite{park2021multiple} designs multiple localized encodes and utilizes component labels as weak supervision to guide each encoder to obtain different local style patterns. In addition, some efforts \cite{li2021few,zhang2022mf,tang2022few} attempt to use an attention mechanism to capture multi-level style patterns.

\begin{figure}[t]
\centering
\includegraphics[width=1.0\linewidth]{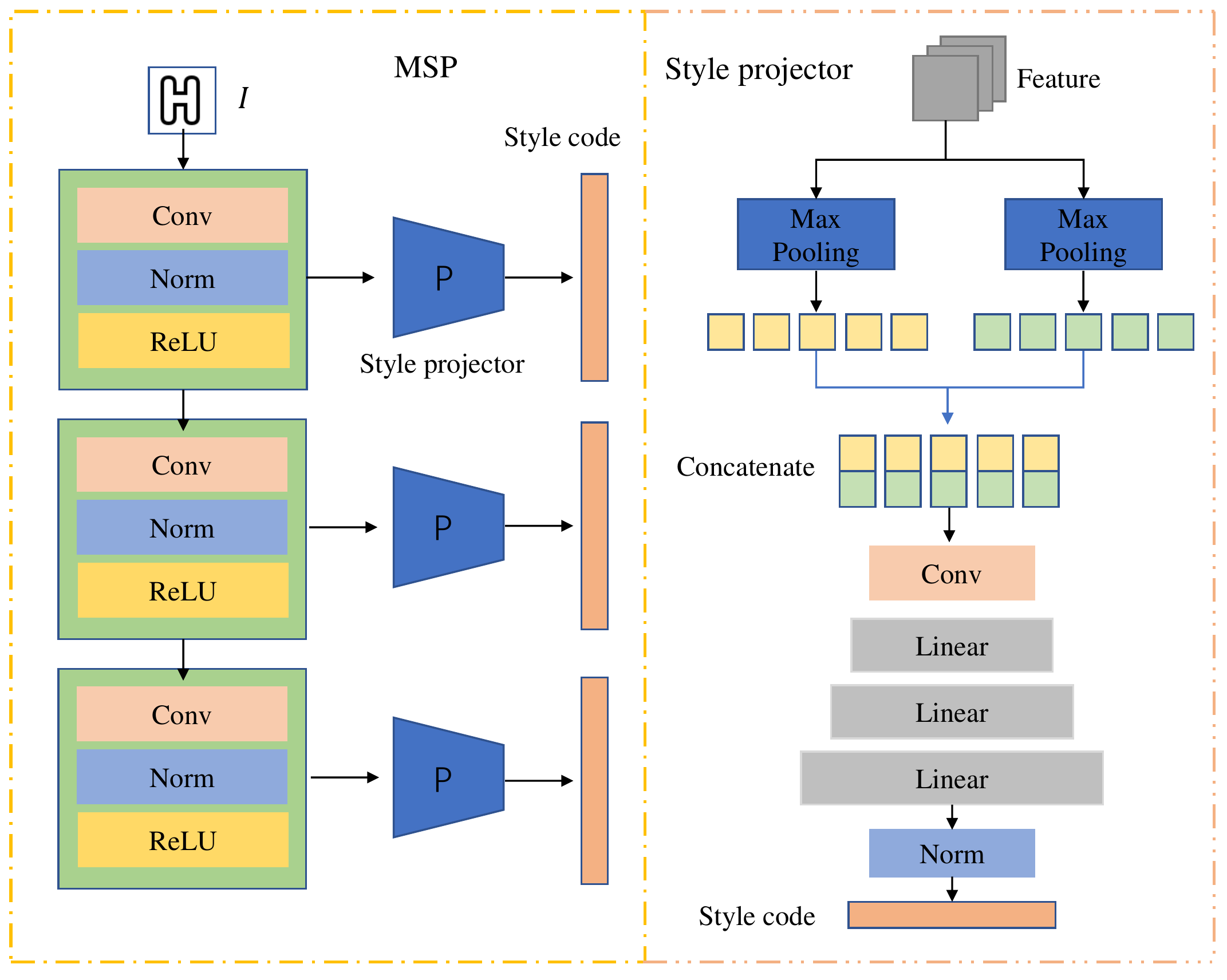}
\caption{\textbf{Overview of MSP module.} When we feed the image into the MSP module, the MSP module first extracts the multi-scale features of the image through convolution, normalization, and activation functions. Then the style projector processes the features through several multi-layer perceptron layers to get the final style codes $\left\{z \right\}$.
 }
\label{fig:meth_MSP}
\end{figure}

\section{Method}
Our model aims to generate the character of the content images with the font of the style images. As shown in Figure \ref{fig:meth_arch}, DS-Font consists of three key components: (1) a multi-layer style projector (MSP); (2) a multi-task patch discriminator; (3) a generator based on an attention mechanism. When training the model, the generator based on the attention mechanism generates images through a content image and a few style images. Then to ensure that a reasonable style representation is obtained, the MSP module projects the generated image and style image into the style code and calculates the cluster-level contrastive style loss. Finally, the multi-task patch discriminator calculates the adversarial loss according to the generated image and ground truth to improve the quality of the generated results.

\subsection{Multi-layer Style Projector}
As mentioned above, the style of the glyphs is complex and fine-grained. When designing the glyphs, experts need to consider multiple levels of styles, $e.g.,$ component-level, stroke-level, and even edge-level. Therefore, a style projector with good performance should extract the features of images at different scales. To this end, we propose an MSP module, as shown in Figure \ref{fig:meth_MSP}. Inspired by the arbitrary image style transfer method  \cite{zhang2022domain}, our MSP module does not use a fusion of multiple layers but projects features of multiple layers into separate latent style spaces to encode different fine-grained styles.

Specifically, we first encode the image to obtain features of different scales (we use three scales, and their proportions to the original image size are 1, 1/2, and 1/4, respectively). Then we obtain the channel-wise mean and peak values of the feature map through average pooling and maximum pooling. These global statistics are closely related to style. Finally, we project features into a set of $K$-dimensional latent style codes $\left\{z_i|i\in[1,3],z_i\in R^{K}\right\} $ through several multilayer perceptron layers. Extracted style code can be denoted as:

\begin{align}
    z_i = M(I_s),   i\in[1,3],
    \label{eq:MSP}
\end{align}%
where $M$ denotes the MSP module and $I_s$ is the style image.

\subsection{Multi-task Patch Discriminator}
The discriminator is designed to distinguish whether the generated image is real or fake. Some previous works \cite{li2021few,zhang2022mf} used two patch discriminators to distinguish the content and style of the generated image, respectively, while others used multi-task discriminators \cite{park2021few,park2021multiple,xie2021dg} to identify different styles or content. However, these are not optimal solutions. The former only identifies the true and false, and does not consider the difference between different categories; The latter distinguish the true and false according to the whole image, ignoring the contribution of each part of the image. We combine the complementary strength of the patch discriminator and multi-task discriminator and propose a novel multi-task patch discriminator. The proposed multi-task patch discriminator considers the different areas of the image when identifying each style (or content) independently. It can not only better distinguish each style (or content) but also make the local details of the generated image more consistent with the ground truth.

 Specifically, we first encode the image by the convolutional blocks with shared weight to obtain features with spatial dimensions. Then two embedding layers are designed according to the type of fonts and the number of characters. The embedding layer projects different styles (or contents) into unique vectors. Finally, by dot multiplication, the image features are combined with the unique vectors to obtain an output matrix with the size of $N \times N$

\subsection{Generator}
Our generator is based on the encoder-transformation-decoder structure. In the design of the transformation module, inspired by \cite{li2021few}, we aggregate style features through self-attention and layer-attention modules to capture local and global features of style images.

\begin{figure*}[t]
\centering
\includegraphics[width=1.0\linewidth]{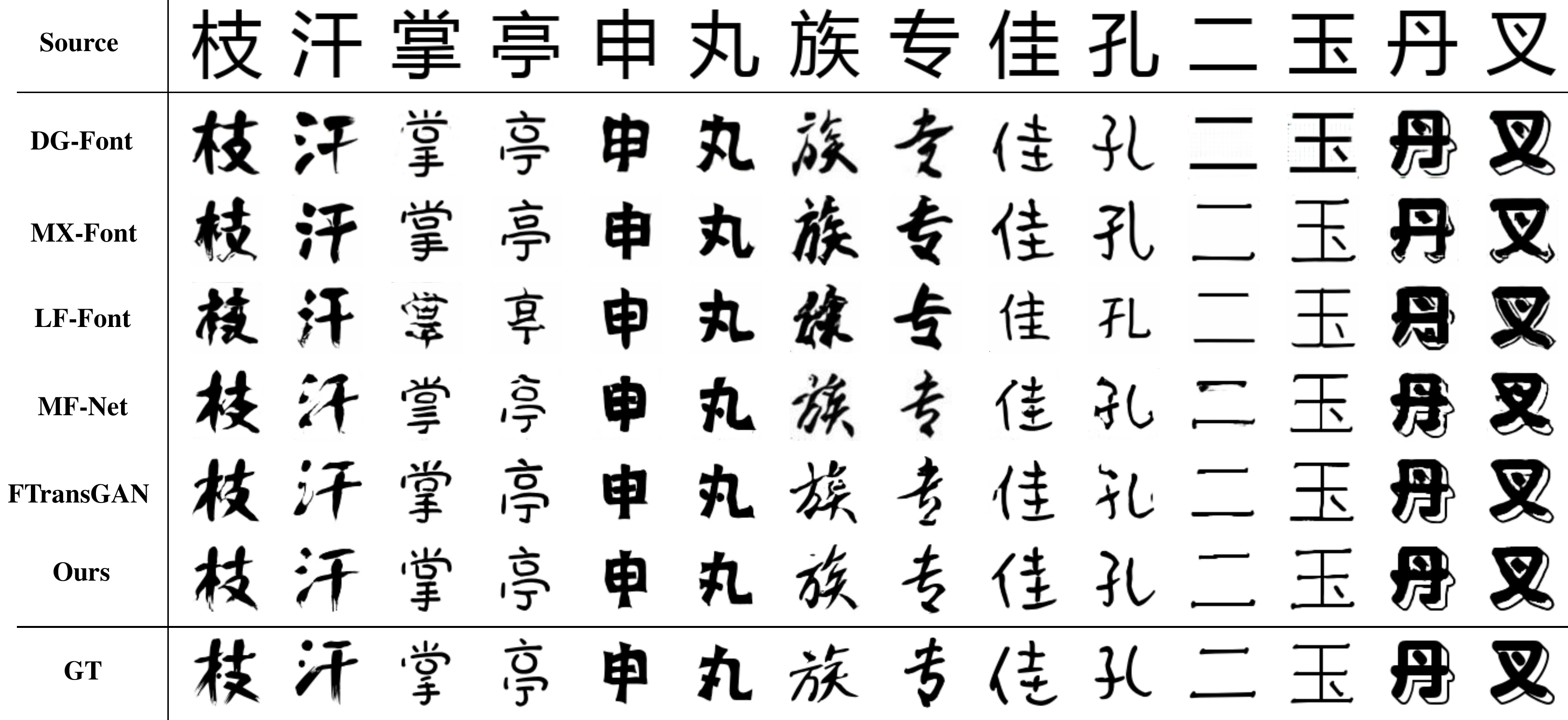}
\caption{\textbf{Qualitative evaluation on the test set.} We show the results generated by DS-Font and five comparison methods. We also provide the source images used for the generation in the top row. The available ground truth images (GT) are shown in the bottom row. 
}
\label{fig:results}
\end{figure*}

\begin{table*}[t]
\centering
\begin{tabular}{cccccccc} 
\toprule
Methods & L1 loss $\downarrow$ & LPIPS $\downarrow$ & RMSE $\downarrow$  & Acc(C) $\uparrow$ & Acc(S) $\uparrow$ & FID(C) $\downarrow$ & FID(S) $\downarrow$ \\
\midrule
\multicolumn{8}{c}{Unseen Characters Seen Fonts}   \\
\midrule 
DG-Font & 0.163 & 0.185 & 0.328 &  \underline{0.97}  & 0.294 &  \underline{10.034} & 51.311 \\
MX-Font & 0.195 & 0.229  & 0.373 & \textbf{0.999} & 0.193 & 10.473 & 60.645 \\
LF-Font & 0.177 & 0.228 & 0.354 & 0.854 & 0.0433 & 18.781 & 76.087 \\
MF-Net & 0.137 & 0.179  & 0.302 & 0.966 & 0.47 & 11.219 & 42.079 \\
FTransGAN & \underline{0.131}  &  \underline{0.166} &  \underline{0.293} & 0.967 &  \underline{0.554} & 10.225 &  \underline{38.52} \\
Ours  & \textbf{0.129} & \textbf{0.159} & \textbf{0.29} & \underline{0.97}  & \textbf{0.807} & \textbf{9.059} & \textbf{23.664} \\
\midrule
\multicolumn{8}{c}{Unseen Fonts Seen Characters} \\
\midrule
DG-Font &  0.175 &  0.202 &0.332 & \underline{0.997} & 0.114 & 7.183 & 47.82 \\
MX-Font & 0.182 & 0.208 & 0.346 & \textbf{0.999} & 0.183 & 7.644 & 50.756 \\
LF-Font & 0.175 & 0.217 & 0.338 & 0.862 &  0.166 & 19.24 & 54.95 \\
MF-Net & 0.135 & 0.16 & 0.283 & \textbf{0.999} & 0.311 & 6.483 & 30.202 \\
FTransGAN & \underline{0.13} & \textbf{0.145}  & \underline{0.272} &  \textbf{0.999} & \underline{0.328} & \textbf{5.76} &  \underline{28.638} \\
Ours  & \textbf{0.129} & \underline{0.146} & \textbf{0.271} &  \textbf{0.999} & \textbf{0.524} &  \underline{5.81} & \textbf{21.119} \\
\bottomrule
\end{tabular}
\caption{\textbf{Quantitative evaluation on the test set.} We evaluate the methods on the test set including UCSF and UFSC set. The bold numbers indicate the best, and underlined numbers represent the second best.}\vspace{-2mm}
\label{tab:resluts}
\end{table*}

\section{Loss Function}
Our goal is to automatically generate high-quality font images via a content image and a few style images. To achieve this goal, we first need to extract a reasonable style representation from the reference image so that the style of the generated image is consistent with the style of the reference image. Secondly, the generated image should be as similar to the ground truth as possible, and there should be no artifacts.

To this end, three losses are proposed to train our model: 1) Cluster-level Contrastive Style (CCS) loss provides proper guidance for the networks to learn the style representation of fonts. 2) Adversarial loss is used to produce realistic images. 3) L1 loss constraints the generated images are similar to the real images. Next, we will introduce each loss in detail in this section. 

\subsection{Cluster-level Contrastive Style Loss}
Contrastive learning (CL) \cite{arora2019theoretical} is based on a surrogate task that treats instances as classes and aims to learn an invariant instance representation. Specifically, contrastive learning aims to compare pairs of image features to draw the positive sample pairs closer and the negative sample pairs apart. Cluster Contrast \cite{dai2022cluster} uses an offline clustering method to build a cluster-level memory dictionary and apply the ideology of momentum update policy from MoCo \cite{he2020momentum,chen2020improved} to the cluster-level memory, which solves the problem of feature inconsistency in memory dictionary and reduces the computational cost. To this end, we adopt contrastive learning and design a new Cluster-level Contrastive Style (CCS) loss to train the MSP module and provide guidance for the generator. In our task, CCS loss is used for pulling the same fonts together and the different fonts away. In addition, we have style (font) labels for supervision and training.

When training the MSP module, at the beginning of each epoch, we encode all the style images $I_s$ used for training through the MSP module and divide them into different classes according to the style (font) labels. (Note $I^+$ denotes the images with the same style as the ground truth, and $I^-$ denotes the images with different styles as the ground truth.) Then we calculate the class center of each class and save it in the memory dictionary. Next, in each iteration, we feed the ground truth (target font) into the MSP module to extract a set of $K$-dimensional vectors ${z}$ as a query $q$. In the memory dictionary, the style vector ${z^+}$ projected by image $I^+$ is used as a positive sample, and the style vector ${z^-}$ projected by image $I^-$ is used as a negative sample. All vectors are normalized to prevent collapse. The contrastive representation learns the visual styles of images by maximizing the mutual information between $q$ and ${z^+}$ in contrast to other style vectors within the memory dictionary considered as negative samples ${z^-}$. Specifically, following \cite{dai2022cluster}, we define the cluster-level contrastive style loss function to train the MSP module as:

\begin{align}
    \mathcal{L}_{ccs}^{MSP} = -\sum_{i=1}^{3}\log\frac{exp(q\cdot z_i^{+}/\tau)}{\sum_{j=1}^{N}exp(q\cdot z_{i_j}/\tau)},
    \label{eq:CluNCE_loss}
\end{align}
where $\cdot$ denotes the dot production, $z_{i_j}$ denotes all samples in the memory dictionary, which includes positive samples $z_i^{+}$ and negative samples $z_i^{-}$ and  $N$ is the number of samples. $\tau$ is a temperature scaling factor and is set to be 0.05 in all our experiments. As for the update of vectors in the memory dictionary, we only use the momentum update strategy to update vectors in each iteration and set the momentum to 0.1.

The contrastive representation also provides proper guidance for the generator to generate high-quality glyphs. We employ the same loss form used for learning MSP in Eq. \ref{eq:CluNCE_loss}. The difference is that the query is not the ground truth but the generated images:

\begin{align}
    \mathcal{L}_{ccs}^{G} = -\sum_{i=1}^{3}\log\frac{exp(\hat{q}\cdot \hat{z}_i^{+}/\tau)}{\sum_{j=1}^{N}exp(\hat{q}\cdot \hat{z}_{i_j}/\tau)},
    \label{eq:G_NCE}
\end{align}
where $\hat{q}$ and $\hat{z}_i^+$ denote the contrastive representation of $I_{cs}$ and $I^+$, respectively. $\hat{z}_{i_j}$ denotes all the samples in memory dictionary.

\subsection{Other Loss Function}
\textbf{Adversarial loss.} Our proposed method is based on Generative Adversarial Networks (GAN). The optimization of GAN is essentially a game problem. Generator G tries to generate a fake image to fool the discriminator D, while the discriminator distinguishes whether the input is real or fake. To generate high-fidelity fonts, we employ the hinge loss functions \cite{miyato2018spectral} as:

\begin{align}
  \mathcal{L}_{adv}^{D} = \mathbf{E}_{(I_r,y_c,y_s)} \left[ [1-D(I_r,y_s)] +[1-D(I_r,y_c)] \right] +   \nonumber\\ \mathbf{E}_{(I_{cs},y_c,y_s)} \left[ [1-D(I_{cs},y_s)] +[1-D(I_{cs},y_c)] \right],
    \label{eq:adv_D}  
\end{align}
    
\begin{align}
    \mathcal{L}_{adv}^{G} = \mathbf{E}_{(I_{cs},y_c,y_s)} \left[ D(I_{cs},y_s) +D(I_{cs},y_c) \right],
    \label{eq:adv_G}
\end{align}
where $I_r$ denotes the ground truth, $y_c$ and $y_s$ denote the content label and style label of the image, respectively.

\textbf{L1 loss.} In oreder to guarantee the pixel-level consistency between generated image $I_{cs}$ and ground truth $I_r$, we employ an L1 loss:

\begin{align}
    \mathcal{L}_{1} = \left \| I_r-I_{cs} \right \|_1.
    \label{eq:L1}
\end{align}

\subsection{Full Objective}
Consequently, we combine all the above loss functions as our full objective to train our model:

\begin{align}
    \mathcal{L}_{D} = \mathcal{L}_{adv}^{D},
    \label{eq:D}
\end{align}

\begin{align}
    \mathcal{L}_{G} = \lambda_1\mathcal{L}_1 +\lambda_2\mathcal{L}_{adv}^{G} + \lambda_3 \mathcal{L}_{ccs}^{G},
    \label{eq:G}
\end{align}

\begin{align}
    \mathcal{L}_{MSP} = \mathcal{L}_{ccs}^{MSP},
    \label{eq:MSP}
\end{align}

\begin{align}
    \mathcal{L} = \mathcal{L}_{D} + \mathcal{L}_{G} +\mathcal{L}_{MSP},
    \label{eq:MSP}
\end{align}
where $\lambda_i$ are hyperparameters to control the weight of each loss function

\section{Experiments}
\subsection{Experimental Settings}
\textbf{Dataset.} The proposed model is trained on the dataset proposed by FTransGAN \cite{li2021few}. The dataset includes 847 gray-scale fonts (style input), each font with approximately 1000 commonly used Chinese characters and 52 Latin letters in the same style. The test set contains two parts: Unseen Characters Seen Fonts (UCSF) and Unseen Fonts Seen Characters (UFSC). They contain 29 characters and fonts as unknown contents and styles and leave the rest part as training data.

\textbf{Training details.} DS-Font is implemented by Pytorch. For experiments, we use Chinese characters as the content and style input. Specifically, we use an ordinary font ($e.g.,$ Microsoft YaHei) as our content input and use 6 Chinese characters as the style input. We set $\lambda_1$ = 100, $\lambda_2$ = 1 and $\lambda_3$ = 0.5 in Eq. \ref{eq:G}. We train DS-Font for 20 epochs using Adam optimizer with a batch size of 256.

\subsection{Comparison With State-of-the-art Methods}
\textbf{Baselines.}
In this subsection, we compare our model with the following methods for Chinese font generation: 1) LF-Font \cite{park2021few} utilizes a factorization strategy to disentangle complex glyph structures and localized style representations. 2) MX-Font \cite{park2021multiple} designs multiple localized experts and utilizes component labels as weak supervision to guide each expert to obtain style representations. 3) DG-Font \cite{xie2021dg} proposes feature deformation skip connection based on the unsupervised method TUNIT \cite{baek2021rethinking}, which enables the model to perform better on cursive characters. 4) FTransGAN \cite{li2021few} specially designs a context-aware attention network to capture the local and global features of style images. 5) MF-Net \cite{zhang2022mf} design a language complexity-aware skip connection to adaptively retain the structural information of the glyph. Note that all methods use Chinese as the input of content and style.

\begin{figure}[t]
\centering
\includegraphics[width=1.0\linewidth]{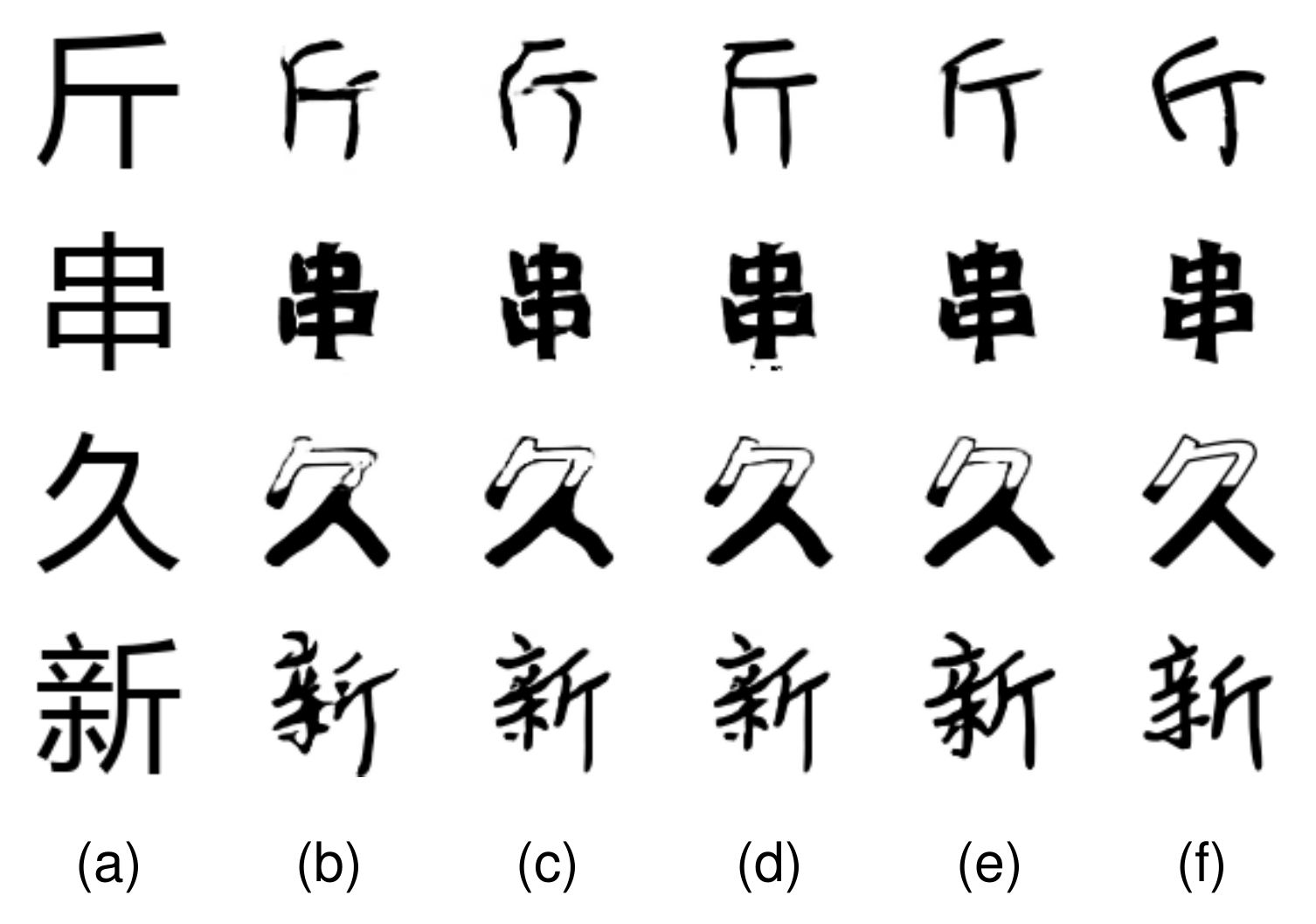}
\caption{\textbf{Visual quality comparison of different variants of our method.} (a) The source image. (b) Using two patch discriminators without CCS loss. (c) Using a multi-task discriminator without CCS loss. (d) Using a multi-task patch discriminator without CCS loss. (e) Using a multi-task patch discriminator with CCS loss. (f) The ground truth images.
}\vspace{-4mm}
\label{fig:ablation}
\end{figure}

\textbf{Evaluation metrics}
To evaluate the performance of our method on FFG tasks, we adopt a variety of similarity metrics from pixel-level to perceptual-level ($e.g.,$ L1 loss, LPIPS). At the perceptual-level, LPIPS measures the distance between the generated image and the ground truth on multi-scale features. In addition, following the previous work, we train two evaluation classifiers for classifying character labels (content awareness) and font labels (style awareness). The detailed settings of the evaluation classifiers are the same as \cite{park2021multiple}. Using classifiers, we measure the classification accuracy of style and content labels. We also calculate Frechét Inception Distance score (FID) scores through style and content classifiers. We denote metrics computed by content and style classifiers as content-aware and style-aware.

\begin{table*}[t]
\centering
\begin{tabular}{cccccccc} 
\toprule
Setting & L1 loss $\downarrow$ & LPIPS $\downarrow$  & RMSE $\downarrow$  & Acc(C) $\uparrow$ & Acc(S) $\uparrow$ & FID(C) $\downarrow$ & FID(S) $\downarrow$ \\
\midrule
\multicolumn{8}{c}{Unseen Characters Seen Fonts}   \\
\midrule 
Patch Dis & 0.131  & 0.166 & 0.293 & \underline{0.967} & 0.554 & 10.225 & 38.52 \\
Multi-task Dis & \underline{0.129} & 0.172 & \underline{0.291}  & 0.957 & 0.583 & \underline{10.003} & 38.185 \\
Our Dis & \textbf{0.128} & \underline{0.165} & \textbf{0.29} & 0.954 & \underline{0.615}   & 10.205 & \underline{34.436} \\
Add CCS loss & \underline{0.129} & \textbf{0.159} & \textbf{0.29} & \textbf{0.97} & \textbf{0.807} & \textbf{9.059} & \textbf{23.664} \\
\midrule
\multicolumn{8}{c}{Unseen Fonts Seen Characters} \\
\midrule
Patch Dis  & 0.13 & \textbf{0.145} & 272 & \textbf{0.999} & 0.328 & \textbf{5.76} & 28.638  \\
Multi-task Dis & \underline{0.127} & \underline{0.146} & \underline{0.269} & \underline{0.997} & 0.365  & 6.143 & 28.823 \\
Our Dis & \textbf{0.126} & 0.147 & \textbf{0.268} & 0.991 & \underline{0.399} & 6.719 & \underline{26.478} \\
Add CCS loss & 0.129 & \underline{0.146} & 0.271 & \textbf{0.999} & \textbf{0.524} & \underline{5.81} & \textbf{21.119} \\
\bottomrule
\end{tabular}
\caption{\textbf{Effect of different components in our method.} The first three lines of two sub-tables evaluate the model trained by using different discriminators without CCS loss, to prove the effectiveness of the discriminator we designed. The last line adds comparative learning based on the proposed Multi-task patch discriminator. The effect of comparative learning can be analyzed by comparing the last two lines.
}\vspace{-2mm}
\label{tab:ablation}
\end{table*}

\textbf{Quantitative comparison.}
Table \ref{tab:resluts} shows the performance of DS-Font and other FFG methods. We report a variety of similarity metrics from pixel-level to perceptual-level and train two evaluation classifiers for classifying character labels and font labels. The values reported in UCSF and UFSC datasets are the average of 29 different characters and fonts, respectively. In the table, we observe that DS-Font outperforms other methods on UCSF and UFSC datasets with most of the evaluation metrics. Specifically, on the UFSC test set, our method is the best in almost all metrics. On the USFC test set, DS-Font is equivalent to FTransGAN in some evaluation metrics (such as L1 loss and LPIPS). It is noted that these metrics focus on the average difference between the generated image and ground truth. They are insensitive to local details and ignore the overall feature distribution similarity between the generated font and the ground truth. In the evaluation metrics FID (S) and Acc (S), we can observe that DS-Font is significantly superior to the comparison method. It demonstrates that our proposed DS-Font can extract a distinctive style representation and generate glyphs consistent with the reference font styles.
 
\textbf{Qualitative comparison.}
In this section, we will quantitatively compare our method with the approaches introduced above. The results are shown in Figure \ref{fig:results}. Our proposed DS-Font can capture the detailed style patterns of the stroke ends, but other methods can not ($e.g.,$ the first two columns). LF-Font \cite{park2021few} cannot maintain the global structure of the original character while generating the target font ($e.g.,$ the third and fourth columns). MX-Font \cite{park2021multiple} fails to generate some complex fonts ($e.g.,$ the last two columns). DG-Font \cite{xie2021dg} introduces the feature deformation skip connection to predict the displacement map, but in some cases, the generated font and the target font are still different in geometry ($e.g.,$ the seventh and eighth columns). FTransGAN \cite{li2021few} proposes a context-aware attention network to capture global and local features of the style images, which ignores fine-grained Local Styles ($e.g.,$ the eleventh and twelfth columns). MF-Net \cite{zhang2022mf} causes artifacts and black spots when generating images ($e.g.,$ the tenth and eleventh columns). As shown in the sixth line, DS-Font captures fine-grained styles and achieves satisfactory results through the proposed discriminator and MSP module with a contrastive learning strategy.

\section{Ablation Study}

\textbf{Multi-task patch discriminator.}
We compare the performances of the patch discriminator, multi-task discriminator, and our multi-task patch discriminator in the FFG task. Note that in the patch discriminator setting, we use two patch discriminators, one to distinguish the content (character) and the other to distinguish the style (font). The results are shown in Figure \ref{fig:ablation} and Table \ref{tab:ablation}. We observe that the multi-task patch discriminator is superior to the other two discriminators in almost metrics except for the content-aware metrics ($e.g.,$ Acc (C)). For the accuracy of the content (Acc (C)), the accuracy of different methods is more than 90 percent, and the gap between them is very small, which has proved that the generated glyphs have strong recognition in content. Meanwhile, as shown in the third line of Figure \ref{fig:ablation}, the model trained under our multi-task patch discriminator maintains the delicate content structure while obtaining the reference image style.

\textbf{Cluster-level contrastive style loss.}
We explore the effect of the CCS loss function by comparing the results of a full model and a model trained without CCS loss. The evaluation results are reported in Table \ref{tab:ablation}. In the last two rows of the table, we find that CCS loss significantly improves the classification accuracy of style and content labels. And from the FID score, we observe that the feature distribution of the image generated by the model with CCS loss training is closer to the real image. In addition, it can be seen from Figure \ref{fig:ablation} that the model with CCS loss function training can better capture local details. According to the above results, we know that CCS loss can guide the MSP module to obtain a reasonable style representation and help the generator capture fine-grained styles.

\section{Conclusion}
In this paper, we propose an effective FFG method by learning the difference between different styles and the similarity of the same style. We design an MSP module for style encoding and propose a cluster-level contrastive style loss function. This loss function guides the MSP module to extract a distinctive style representation and help train the generator. We further propose a multi-task patch discriminator, which considers the contribution of each region of the image when distinguishing each style independently. Extensive experimental results demonstrate that our DS-Font achieves superior font generation results compared with state-of-the-art methods.

\small
\bibliographystyle{named}
\bibliography{DS-Font/DS-Font3}

\begin{thebibliography}{}

\bibitem[\protect\citeauthoryear{Anokhin \bgroup \em et al.\egroup
  }{2020}]{anokhin2020high}
Ivan Anokhin, Pavel Solovev, Denis Korzhenkov, Alexey Kharlamov, Taras
  Khakhulin, Aleksei Silvestrov, Sergey Nikolenko, Victor Lempitsky, and Gleb
  Sterkin.
\newblock High-resolution daytime translation without domain labels.
\newblock In {\em Proceedings of the IEEE/CVF Conference on Computer Vision and
  Pattern Recognition}, pages 7488--7497, 2020.

\bibitem[\protect\citeauthoryear{Arora \bgroup \em et al.\egroup
  }{2019}]{arora2019theoretical}
Sanjeev Arora, Hrishikesh Khandeparkar, Mikhail Khodak, Orestis Plevrakis, and
  Nikunj Saunshi.
\newblock A theoretical analysis of contrastive unsupervised representation
  learning.
\newblock {\em arXiv preprint arXiv:1902.09229}, 2019.

\bibitem[\protect\citeauthoryear{Baek \bgroup \em et al.\egroup
  }{2021}]{baek2021rethinking}
Kyungjune Baek, Yunjey Choi, Youngjung Uh, Jaejun Yoo, and Hyunjung Shim.
\newblock Rethinking the truly unsupervised image-to-image translation.
\newblock In {\em Proceedings of the IEEE/CVF International Conference on
  Computer Vision}, pages 14154--14163, 2021.

\bibitem[\protect\citeauthoryear{Cha \bgroup \em et al.\egroup
  }{2020}]{cha2020few}
Junbum Cha, Sanghyuk Chun, Gayoung Lee, Bado Lee, Seonghyeon Kim, and Hwalsuk
  Lee.
\newblock Few-shot compositional font generation with dual memory.
\newblock In {\em European Conference on Computer Vision}, pages 735--751.
  Springer, 2020.

\bibitem[\protect\citeauthoryear{Chang \bgroup \em et al.\egroup
  }{2018}]{chang2018chinese}
Jie Chang, Yujun Gu, Ya~Zhang, Yan-Feng Wang, and CM~Innovation.
\newblock Chinese handwriting imitation with hierarchical generative
  adversarial network.
\newblock In {\em BMVC}, page 290, 2018.

\bibitem[\protect\citeauthoryear{Chen \bgroup \em et al.\egroup
  }{2020}]{chen2020improved}
Xinlei Chen, Haoqi Fan, Ross Girshick, and Kaiming He.
\newblock Improved baselines with momentum contrastive learning.
\newblock {\em arXiv preprint arXiv:2003.04297}, 2020.

\bibitem[\protect\citeauthoryear{Choi \bgroup \em et al.\egroup
  }{2020}]{choi2020stargan}
Yunjey Choi, Youngjung Uh, Jaejun Yoo, and Jung-Woo Ha.
\newblock Stargan v2: Diverse image synthesis for multiple domains.
\newblock In {\em Proceedings of the IEEE/CVF conference on computer vision and
  pattern recognition}, pages 8188--8197, 2020.

\bibitem[\protect\citeauthoryear{Dai \bgroup \em et al.\egroup
  }{2022}]{dai2022cluster}
Zuozhuo Dai, Guangyuan Wang, Weihao Yuan, Siyu Zhu, and Ping Tan.
\newblock Cluster contrast for unsupervised person re-identification.
\newblock In {\em Proceedings of the Asian Conference on Computer Vision},
  pages 1142--1160, 2022.

\bibitem[\protect\citeauthoryear{Gao and Wu}{2020}]{gao2020gan}
Yiming Gao and Jiangqin Wu.
\newblock Gan-based unpaired chinese character image translation via skeleton
  transformation and stroke rendering.
\newblock In {\em proceedings of the AAAI conference on artificial
  intelligence}, volume~34, pages 646--653, 2020.

\bibitem[\protect\citeauthoryear{Gao \bgroup \em et al.\egroup
  }{2019}]{gao2019artistic}
Yue Gao, Yuan Guo, Zhouhui Lian, Yingmin Tang, and Jianguo Xiao.
\newblock Artistic glyph image synthesis via one-stage few-shot learning.
\newblock {\em ACM Transactions on Graphics (TOG)}, 38(6):1--12, 2019.

\bibitem[\protect\citeauthoryear{Goodfellow \bgroup \em et al.\egroup
  }{2020}]{goodfellow2020generative}
Ian Goodfellow, Jean Pouget-Abadie, Mehdi Mirza, Bing Xu, David Warde-Farley,
  Sherjil Ozair, Aaron Courville, and Yoshua Bengio.
\newblock Generative adversarial networks.
\newblock {\em Communications of the ACM}, 63(11):139--144, 2020.

\bibitem[\protect\citeauthoryear{Hassan \bgroup \em et al.\egroup
  }{2021}]{hassan2021unpaired}
Ammar~Ul Hassan, Hammad Ahmed, and Jaeyoung Choi.
\newblock Unpaired font family synthesis using conditional generative
  adversarial networks.
\newblock {\em Knowledge-Based Systems}, 229:107304, 2021.

\bibitem[\protect\citeauthoryear{He \bgroup \em et al.\egroup
  }{2020}]{he2020momentum}
Kaiming He, Haoqi Fan, Yuxin Wu, Saining Xie, and Ross Girshick.
\newblock Momentum contrast for unsupervised visual representation learning.
\newblock In {\em Proceedings of the IEEE/CVF conference on computer vision and
  pattern recognition}, pages 9729--9738, 2020.

\bibitem[\protect\citeauthoryear{Huang and Belongie}{2017}]{huang2017arbitrary}
Xun Huang and Serge Belongie.
\newblock Arbitrary style transfer in real-time with adaptive instance
  normalization.
\newblock In {\em Proceedings of the IEEE international conference on computer
  vision}, pages 1501--1510, 2017.

\bibitem[\protect\citeauthoryear{Isola \bgroup \em et al.\egroup
  }{2017}]{isola2017image}
Phillip Isola, Jun-Yan Zhu, Tinghui Zhou, and Alexei~A Efros.
\newblock Image-to-image translation with conditional adversarial networks.
\newblock In {\em Proceedings of the IEEE conference on computer vision and
  pattern recognition}, pages 1125--1134, 2017.

\bibitem[\protect\citeauthoryear{Jiang \bgroup \em et al.\egroup
  }{2017}]{jiang2017dcfont}
Yue Jiang, Zhouhui Lian, Yingmin Tang, and Jianguo Xiao.
\newblock Dcfont: an end-to-end deep chinese font generation system.
\newblock In {\em SIGGRAPH Asia 2017 Technical Briefs}, pages 1--4. 2017.

\bibitem[\protect\citeauthoryear{Jiang \bgroup \em et al.\egroup
  }{2019}]{jiang2019scfont}
Yue Jiang, Zhouhui Lian, Yingmin Tang, and Jianguo Xiao.
\newblock Scfont: Structure-guided chinese font generation via deep stacked
  networks.
\newblock In {\em Proceedings of the AAAI conference on artificial
  intelligence}, volume~33, pages 4015--4022, 2019.

\bibitem[\protect\citeauthoryear{Li \bgroup \em et al.\egroup
  }{2021}]{li2021few}
Chenhao Li, Yuta Taniguchi, Min Lu, and Shin'ichi Konomi.
\newblock Few-shot font style transfer between different languages.
\newblock In {\em Proceedings of the IEEE/CVF Winter Conference on Applications
  of Computer Vision}, pages 433--442, 2021.

\bibitem[\protect\citeauthoryear{Liu \bgroup \em et al.\egroup
  }{2019}]{liu2019few}
Ming-Yu Liu, Xun Huang, Arun Mallya, Tero Karras, Timo Aila, Jaakko Lehtinen,
  and Jan Kautz.
\newblock Few-shot unsupervised image-to-image translation.
\newblock In {\em Proceedings of the IEEE/CVF international conference on
  computer vision}, pages 10551--10560, 2019.

\bibitem[\protect\citeauthoryear{Mirza and
  Osindero}{2014}]{mirza2014conditional}
Mehdi Mirza and Simon Osindero.
\newblock Conditional generative adversarial nets.
\newblock {\em arXiv preprint arXiv:1411.1784}, 2014.

\bibitem[\protect\citeauthoryear{Miyato \bgroup \em et al.\egroup
  }{2018}]{miyato2018spectral}
Takeru Miyato, Toshiki Kataoka, Masanori Koyama, and Yuichi Yoshida.
\newblock Spectral normalization for generative adversarial networks.
\newblock {\em arXiv preprint arXiv:1802.05957}, 2018.

\bibitem[\protect\citeauthoryear{Park \bgroup \em et al.\egroup
  }{2021a}]{park2021few}
Song Park, Sanghyuk Chun, Junbum Cha, Bado Lee, and Hyunjung Shim.
\newblock Few-shot font generation with localized style representations and
  factorization.
\newblock In {\em Proceedings of the AAAI Conference on Artificial
  Intelligence}, volume~35, pages 2393--2402, 2021.

\bibitem[\protect\citeauthoryear{Park \bgroup \em et al.\egroup
  }{2021b}]{park2021multiple}
Song Park, Sanghyuk Chun, Junbum Cha, Bado Lee, and Hyunjung Shim.
\newblock Multiple heads are better than one: Few-shot font generation with
  multiple localized experts.
\newblock In {\em Proceedings of the IEEE/CVF International Conference on
  Computer Vision}, pages 13900--13909, 2021.

\bibitem[\protect\citeauthoryear{Sun \bgroup \em et al.\egroup
  }{2018}]{sun2018pyramid}
Donghui Sun, Qing Zhang, and Jun Yang.
\newblock Pyramid embedded generative adversarial network for automated font
  generation.
\newblock In {\em 2018 24th International Conference on Pattern Recognition
  (ICPR)}, pages 976--981. IEEE, 2018.

\bibitem[\protect\citeauthoryear{Tang \bgroup \em et al.\egroup
  }{2022}]{tang2022few}
Licheng Tang, Yiyang Cai, Jiaming Liu, Zhibin Hong, Mingming Gong, Minhu Fan,
  Junyu Han, Jingtuo Liu, Errui Ding, and Jingdong Wang.
\newblock Few-shot font generation by learning fine-grained local styles.
\newblock In {\em Proceedings of the IEEE/CVF Conference on Computer Vision and
  Pattern Recognition}, pages 7895--7904, 2022.

\bibitem[\protect\citeauthoryear{Tian}{2017}]{tian2017zi2zi}
Yuchen Tian.
\newblock zi2zi: Master chinese calligraphy with conditional adversarial
  networks.
\newblock {\em Internet] https://github. com/kaonashi-tyc/zi2zi}, 3, 2017.

\bibitem[\protect\citeauthoryear{Wen \bgroup \em et al.\egroup
  }{2021}]{wen2021zigan}
Qi~Wen, Shuang Li, Bingfeng Han, and Yi~Yuan.
\newblock Zigan: Fine-grained chinese calligraphy font generation via a
  few-shot style transfer approach.
\newblock In {\em Proceedings of the 29th ACM International Conference on
  Multimedia}, pages 621--629, 2021.

\bibitem[\protect\citeauthoryear{Xie \bgroup \em et al.\egroup
  }{2021}]{xie2021dg}
Yangchen Xie, Xinyuan Chen, Li~Sun, and Yue Lu.
\newblock Dg-font: Deformable generative networks for unsupervised font
  generation.
\newblock In {\em Proceedings of the IEEE/CVF Conference on Computer Vision and
  Pattern Recognition}, pages 5130--5140, 2021.

\bibitem[\protect\citeauthoryear{Yu \bgroup \em et al.\egroup
  }{2019}]{yu2019multi}
Xiaoming Yu, Yuanqi Chen, Shan Liu, Thomas Li, and Ge~Li.
\newblock Multi-mapping image-to-image translation via learning
  disentanglement.
\newblock {\em Advances in Neural Information Processing Systems}, 32, 2019.

\bibitem[\protect\citeauthoryear{Zhang \bgroup \em et al.\egroup
  }{2018}]{zhang2018separating}
Yexun Zhang, Ya~Zhang, and Wenbin Cai.
\newblock Separating style and content for generalized style transfer.
\newblock In {\em Proceedings of the IEEE conference on computer vision and
  pattern recognition}, pages 8447--8455, 2018.

\bibitem[\protect\citeauthoryear{Zhang \bgroup \em et al.\egroup
  }{2019a}]{zhang2019deep}
Mingjin Zhang, Nannan Wang, Yunsong Li, and Xinbo Gao.
\newblock Deep latent low-rank representation for face sketch synthesis.
\newblock {\em IEEE transactions on neural networks and learning systems},
  30(10):3109--3123, 2019.

\bibitem[\protect\citeauthoryear{Zhang \bgroup \em et al.\egroup
  }{2019b}]{zhang2019neural}
Mingjin Zhang, Nannan Wang, Yunsong Li, and Xinbo Gao.
\newblock Neural probabilistic graphical model for face sketch synthesis.
\newblock {\em IEEE transactions on neural networks and learning systems},
  31(7):2623--2637, 2019.

\bibitem[\protect\citeauthoryear{Zhang \bgroup \em et al.\egroup
  }{2022a}]{zhang2022mf}
Yufan Zhang, Junkai Man, and Peng Sun.
\newblock Mf-net: A novel few-shot stylized multilingual font generation
  method.
\newblock In {\em Proceedings of the 30th ACM International Conference on
  Multimedia}, pages 2088--2096, 2022.

\bibitem[\protect\citeauthoryear{Zhang \bgroup \em et al.\egroup
  }{2022b}]{zhang2022domain}
Yuxin Zhang, Fan Tang, Weiming Dong, Haibin Huang, Chongyang Ma, Tong-Yee Lee,
  and Changsheng Xu.
\newblock Domain enhanced arbitrary image style transfer via contrastive
  learning.
\newblock {\em arXiv preprint arXiv:2205.09542}, 2022.

\bibitem[\protect\citeauthoryear{Zhu \bgroup \em et al.\egroup
  }{2017a}]{zhu2017unpaired}
Jun-Yan Zhu, Taesung Park, Phillip Isola, and Alexei~A Efros.
\newblock Unpaired image-to-image translation using cycle-consistent
  adversarial networks.
\newblock In {\em Proceedings of the IEEE international conference on computer
  vision}, pages 2223--2232, 2017.

\bibitem[\protect\citeauthoryear{Zhu \bgroup \em et al.\egroup
  }{2017b}]{zhu2017deep}
Mingrui Zhu, Nannan Wang, Xinbo Gao, and Jie Li.
\newblock Deep graphical feature learning for face sketch synthesis.
\newblock In {\em Proceedings of the 26th international joint conference on
  artificial intelligence}, pages 3574--3580, 2017.

\end{thebibliography}

\normalsize
\clearpage
\appendixtitleon
\appendixtitletocon
\begin{appendices}
\section{Implementation Details}
\textbf{Multi-task patch discriminator.}
As shown in Table \ref{tab:discriminator}, the architecture of the multi-task patch discriminator consists of an encoder with shared weights, a content embedding layer, a style embedding layer, and a correspondence layer. Specifically, the encoder with shared weight encodes the image to obtain features with spatial dimensions. Then two embedding layers project different styles (or contents) into unique vectors. Finally, according to the correspondence layer, the image features are combined with the unique vectors to obtain an output matrix with the size of $N \times N$.

\begin{table}[htbp]
\centering
\begin{tabular}{ccc} 
\toprule
Module &  \multicolumn{1}{|c|}{Layers} & output shape  \\
\midrule
\multirow{5}{*}{Encoder}  & \multicolumn{1}{|c|}{Convblock } & 32 $\times$ 32 $\times$ 32   \\
\multicolumn{1}{c}{} & \multicolumn{1}{|c|}{Resblock } & 16 $\times$ 16 $\times$ 64  \\
\multicolumn{1}{c}{} & \multicolumn{1}{|c|}{Resblock } & 8 $\times$ 8 $\times$ 128  \\
\multicolumn{1}{c}{} & \multicolumn{1}{|c|}{Resblock } & 4 $\times$ 4 $\times$ 256  \\ 
\multicolumn{1}{c}{} & \multicolumn{1}{|c|}{Resblock $\times$ 2} & 4 $\times$ 4 $\times$ 256  \\
\midrule
\multicolumn{1}{c}{Content embedding} & \multicolumn{1}{|c|}{Linear} & 1 $\times$ 1 $\times$ 256 \\
\midrule
\multicolumn{1}{c}{Style embedding} & \multicolumn{1}{|c|}{Linear} & 1 $\times$ 1 $\times$ 256 \\
\midrule
\multicolumn{1}{c}{Correspondence} & \multicolumn{1}{|c|}{Dot product} & 4 $\times$ 4 $\times$ 256 \\
\bottomrule
\end{tabular}
\caption{\textbf{Architecture of multi-task patch discriminator.} k3s1 indicates the convolutional layer with kernel size 3 and stride 1.}
\label{tab:discriminator}
\end{table}

\begin{figure}[t]
\centering
\includegraphics[width=1.0\linewidth]{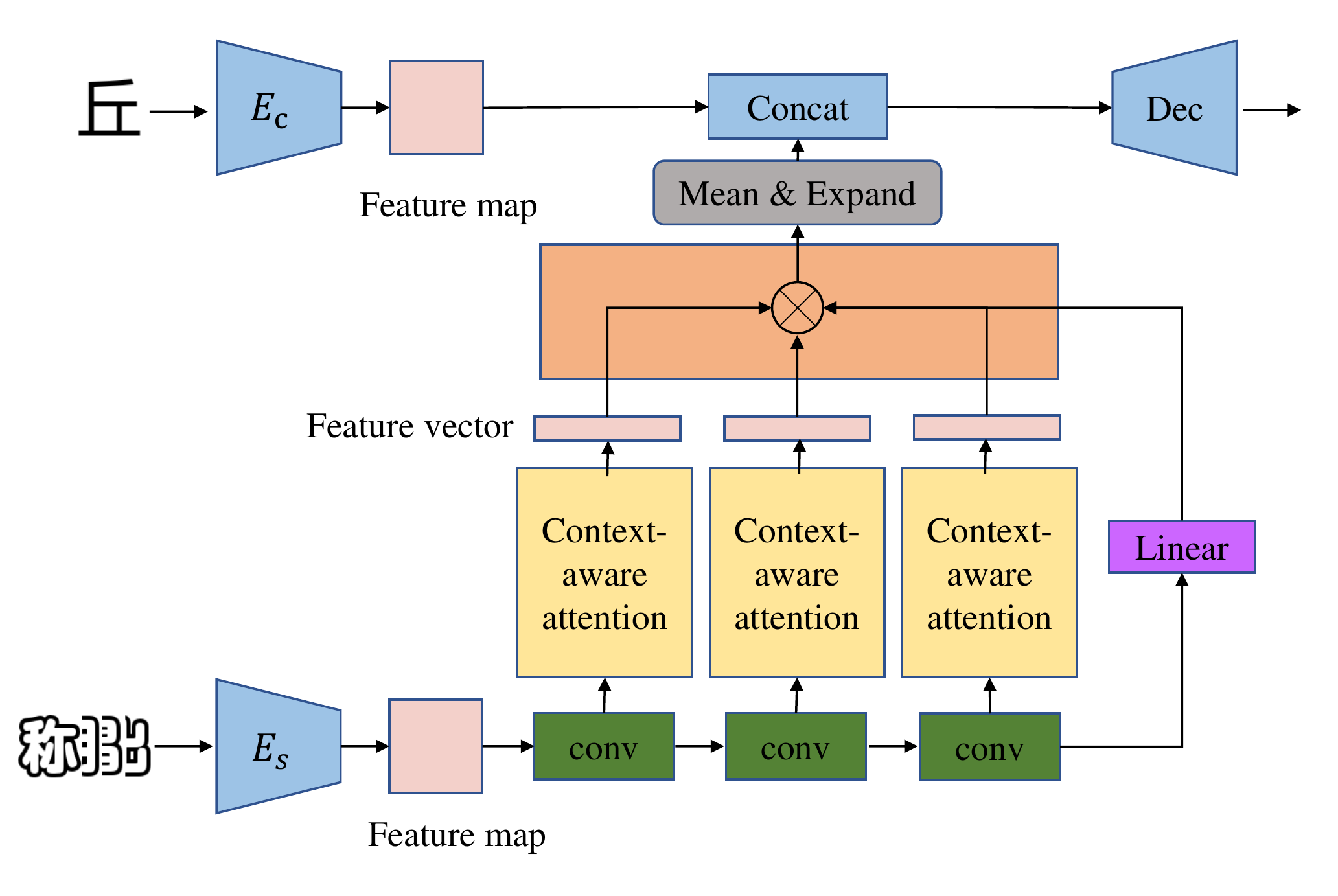}
\caption{\textbf{Overview of the generator.}}
\label{fig:generator}
\end{figure}

\begin{figure}[t]
\centering
\includegraphics[width=1.0\linewidth]{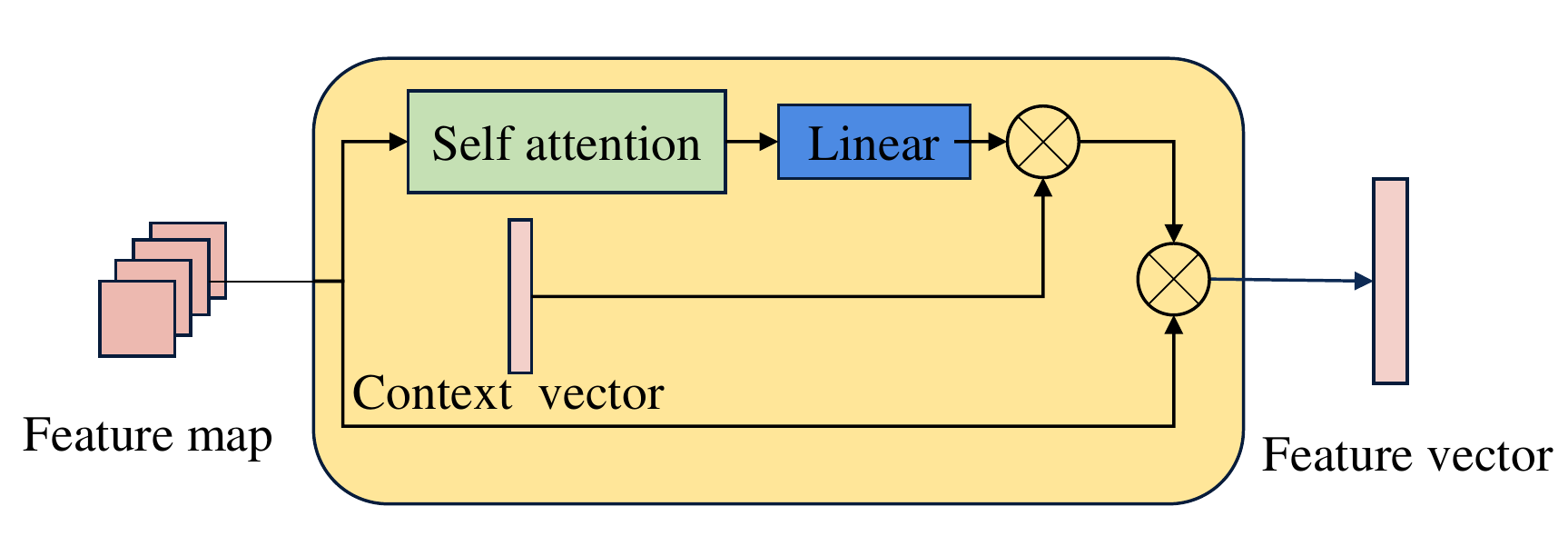}
\caption{\textbf{Architecture of the context-aware attention module.}}
\label{fig:context-aware attention}
\end{figure}

\textbf{Generator.}
The architecture of the generator follows the one used in FTransGAN, which
uses a context-aware attention module and a layer attention module to capture both local and global style features. The overall architecture is shown in Figure \ref{fig:generator} and the context-aware attention module is shown in Figure \ref{fig:context-aware attention}.

\textbf{Training procedure of DS-Font.}
 Our proposed DS-Font consists of three key components: (1) a multi-layer style projector (MSP); (2) a multi-task patch discriminator; (3) a generator based on an attention mechanism. We have already introduced the loss function related to each module. Here, we will briefly introduce the process of training the whole model in Algorithm \ref{alg:algorithm}.
 
\textbf{Code.}
we provide the full code of our proposed DS-Font. Please refer to the ``README.md" file located in the Code folder for detailed usage of the code.

\begin{algorithm}[tb]
    \caption{Training procedure of DS-Font}
    \label{alg:algorithm}
    \textbf{Input}: content images $I_c$, style images $I_s$, ground truth $I_r$\\
    \textbf{Model}: generator $G$, multi-task patch discriminator $D$, multi-layer style projector (MSP) $M$ \\
    \textbf{Output}: generated image $I_{cs}$
    \begin{algorithmic}[1] 
        \FOR{$n$ in [1,$num\_epochs$]}
        \STATE{Extract feature vectors $z$ from $I_s$ by $M$}
        \STATE{Clustering $z$ into K clusters and calculate the cluster centers }
        \STATE{Save cluster centers $z_j$ to the memory dictionary}
        \FOR{$i$ in [1,$num\_iters$]}
        \STATE{Generating $I_{cs}$ from $I_c$ and $I_s$ by $G$}
        \STATE {Compute adversarial loss with Eq. 5}
        \STATE{Update the discriminator $D$ by optimizer}
        \STATE{Compute CCS loss with Eq. 2 }
        \STATE{Update MSP module $M$ by optimizer}
        \STATE{Update cluster centers with momentum update strategy}
        \STATE{Compute loss functions with Eq. 8}
        \STATE{Update the generator $G$ by optimizer}
        \ENDFOR
        \ENDFOR

    \end{algorithmic}
\end{algorithm}

\section{More Font Generation Results}
\begin{figure*}[t]
\centering
\includegraphics[width=1.0\linewidth]{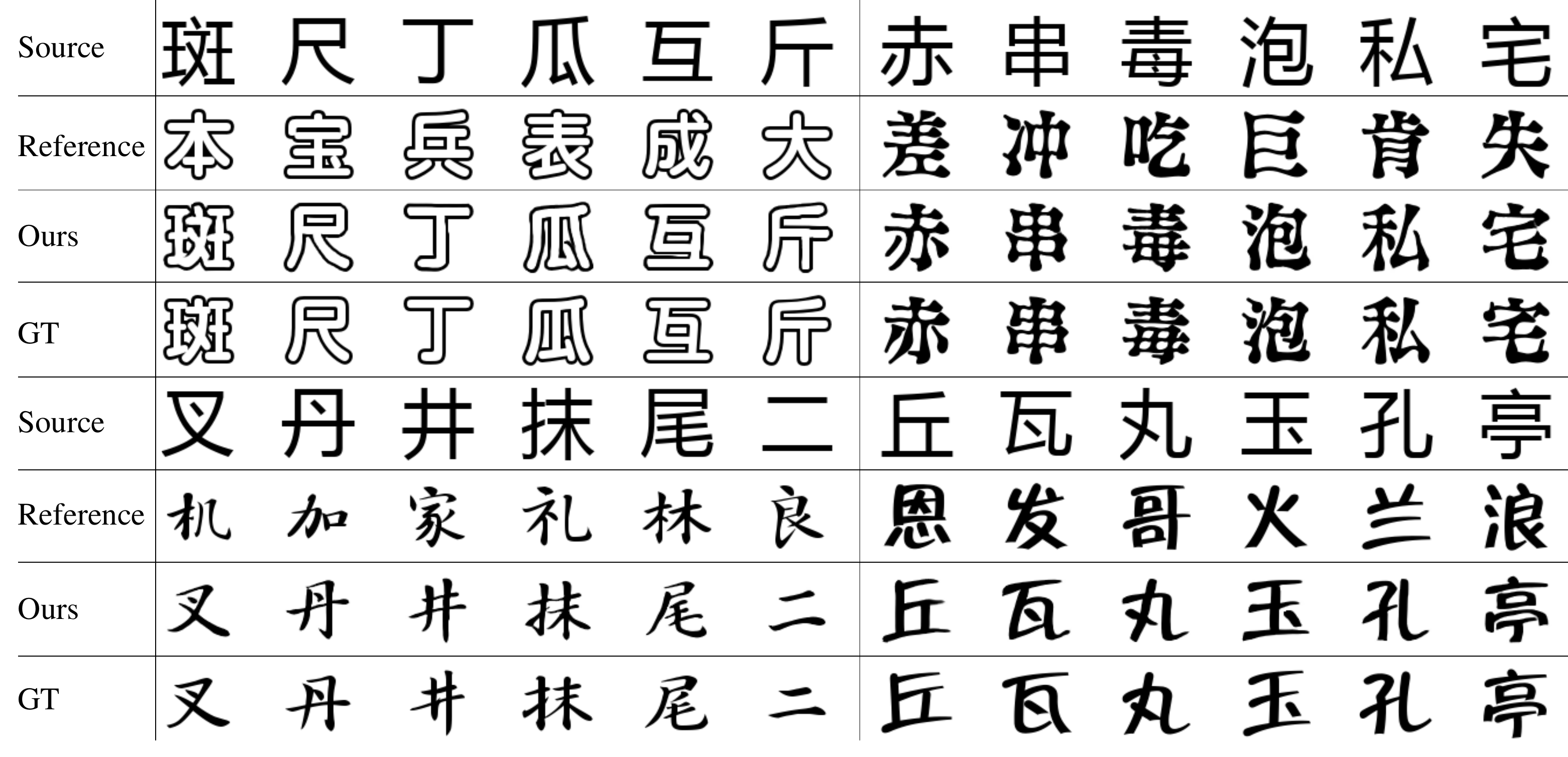}
\caption{\textbf{Font generation results on UCSF set}
}
\label{fig:results1}
\end{figure*}

\begin{figure*}[t]
\centering
\includegraphics[width=1.0\linewidth]{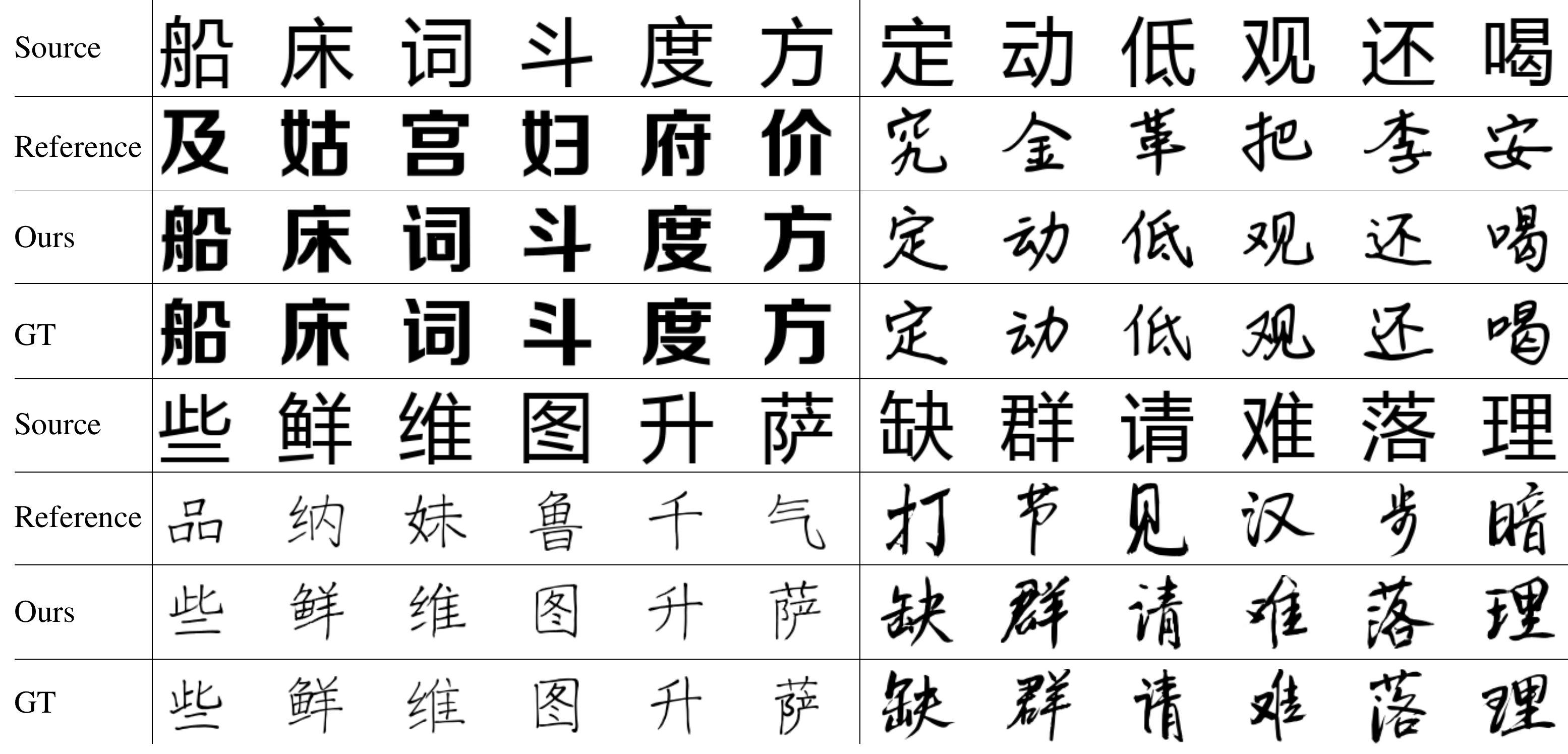}
\caption{\textbf{Font generation results on UFSC set}
}
\label{fig:results2}
\end{figure*}

In Figure \ref{fig:results1} and Figure \ref{fig:results2}, we show more experimental results on the Unseen Character Seen Font (UCSF) and Unseen Font Seen Character (UFSC) set. The results show that DS-Font can capture fine-grained styles and achieve satisfactory results on both USFC and UCSF sets.

\end{appendices}
	
\end{document}